# Robust Pose Invariant Shape and Texture based Hand Recognition

F. Sohel, A. El-Sallam, and M. Bennamoun

*Abstract*— This paper presents a novel personal identification and verification system using information extracted from the hand shape and texture. The system has two major constituent modules: a fully automatic and robust peg free segmentation and pose normalisation module, and a recognition module. In the first module, the hand is segmented from its background using a thresholding technique based on Otsu's method combined with a skin colour detector. A set of fully automatic algorithms are then proposed to segment the palm and fingers. In these algorithms, the skeleton and the contour of the hand and fingers are estimated and used to determine the global pose of the hand and the pose of each individual finger. Finally the palm and fingers are cropped, pose corrected and normalised. In the recognition module, various shape and texture based features are extracted and used for matching purposes. The modified Hausdorff distance, the Iterative Closest Point (ICP) and Independent Component Analysis (ICA) algorithms are used for shape and texture features of the fingers. For the palmprints, we use the Discrete Cosine Transform (DCT), directional line features and ICA. Recognition (identification and verification) tests were performed using fusion strategies based on the similarity scores of the fingers and the palm. Experimental results show that the proposed system exhibits a superior performance over existing systems with an accuracy of over 98% for hand identification and verification (at equal error rate) in a database of 560 different subjects.

*Index Terms*— Biometrics, Identification of persons, image segmentation.

## I. INTRODUCTION

RECENTLY there has been an increased interest in the development of robust and fully automatic recognition systems. This trend led to intensive research in biometrics such as fingerprints [1], face [2, 3], ear [4], iris [5], hand geometry [6, 7], and palmprint recognition [8]. Each biometric has its own strengths and weaknesses depending on the requirements of the intended application as reported in many research surveys [9]. In this paper, the term "recognition" refers to both "identification" and "verification". "Identification" refers to the identification of an individual based on the comparison of collected biometrics with previously acquired (during the enrollment process) biometrics, which have been saved in a database. On the other hand, "verification" refers to the verification process on whether an individual is the person that they claim to be, based upon the validation of a biometric sample collected against a previously collected sample of the individual during enrollment. Therefore, identification is a *"one to many"* matching problem while verification is a *"one to one"* matching. In this work we focus on the development of a fully automatic hand-based recognition system for the following reasons: (i) unlike other biometrics, e.g. iris, the shape of the hand can easily be captured in a relatively user friendly manner using conventional Charge Coupled Device (CCD) cameras (ease of use), (ii) this technology is more acceptable by the society because of its loose connection to forensic applications (acceptability of hand-based biometrics), (iii) relatively little work has been done in this area and, (iv) finally, the performance of multimodal biometric systems (e.g., shape + texture modalities) is superior to the performance of conventional unimodal systems. Furthermore, Multimodal systems are also harder to forge [10].

Most of the hand-based biometric systems which are available in the literature are based on the geometric features of the hand. One of the earliest papers was published by Sanchez-Reillo et al. [11]. Their work selects 25 features, such as finger widths at different latitudes, finger and palm heights, finger deviations and the angles of the inter finger valleys. However, this system is peg-based and uses six tops placed in pre-determined positions to guide the placement of the hand. Peg-based systems can cause users' discomfort and can represent a considerable source of failure in some cases especially in cases where the size of the peg is bigger or smaller than the size of a person's hand [12]. In addition, the system was tested on a small sample size of 20 subjects and achieved a recognition rate of up to 97%. Another major concurrent work was published by Jain et al. in [13]. Their work also uses a peg-based data acquisition system. It requires a controlled background and is only designed for black and white images. They extracted 16 features, which include the length and the width of the fingers, the aspect ratio of the palm to fingers, and the thickness of the hand. Their system was tested in a verification experimental setup for web access for a group of 10 people [14]. Moreover, neither of the two aforementioned systems is scale invariant [13]. Kumar and Zhang [8, 15, 16] successfully used both geometric shape features and texture features. Additional features such as finger

This research was supported by an ARC project (DE120102960).

F. Sohel was with School of Computer Science and Software Engineering, The University of Western Australia, Australia. 35 Stirling Highway Crawley, WA 6009, Australia.Phone: +61 8 6488 2796; Email: Ferdous.Sohel@uwa.edu.au.

A. El-Sallam was with School of Computer Science and Software Engineering, The University of Western Australia, Australia.

M. Bennamoun was with School of Computer Science and Software Engineering, The University of Western Australia, Australia.

widths at various positions and palm size, finger shapes were also used in [17]. However, the size of the palm is highly dependent on the pose of the thumb. Yörük et al. [7][18] presented another major work which primarily used shape based features. Upon the segmentation of the fingers and the palm, it reconstructs the hand based on a predefined template. The reconstructed hand is then used in feature extraction and matching. Our previous work in [6] presented a shape-based hand recognition technique which used geometrical and statistical features. Amayeh et al. [12] present one of the most recent works in this area based on high-order Zernike moments. However, a major drawback of high-order Zernike moments is their sensitivity to even slight changes in the extracted silhouette of the hand [11]. To overcome the aforementioned limitations, we present in this paper a novel automatic hand (shape and texture) based a personal verification and identification system that is both template and peg free. It is robust to both hand and finger poses (i.e., position and orientation) as well as the colour contrast between the foreground and background.

The system has two major constituent modules: a segmentation and pose normalisation module, and a recognition module. The first module introduces a novel hand palm and fingers segmentation algorithm that is accurate, robust to scale, noise, and aspect ratio variations. The algorithm is also invariant to hand and/or finger(s) pose(s) and can determine whether the hand is the right or left hand. In the second module, features are extracted and matched using a range of feature extraction and matching techniques. The score level information from the different matching techniques of the hand (e.g., fingers and the palm) is then fused for the identification and verification of subjects. To segment the hand from the background, we propose an approach combining a thresholding algorithm based on Otsu's method [17] with a skin colour detector (a detailed description is presented in Section III-A). This is followed by a number of image processing steps (e.g., image morphing) to remove isolated pixels and fill in the holes of the segmented hand. Once this is achieved, we extract the skeleton and the contour of the segmented hand to estimate its global pose and the pose of each finger. The palm and the fingers are then cropped and represented in a pre-defined and consistent orientation. Besides being peg-free, our algorithm has several advantages compared with existing techniques. They include: (i) the arm or any object attached to it (e.g., clothes and watches) does not have any effect on the performance of our algorithm, (ii) unlike [12], our algorithm does not require any initialization (e.g., palm radius), (iii) it does not involve an iterative process, but uses simple mathematical calculations, (iv) it is able to automatically exclude non regions of interests such as the arm portion with or without clothing or objects (e.g. watches) which has been a dominant problem in many existing hand biometrics systems that require manual interventions, (v) existing methods restrict the maximum pose angle of the hand to be less than 45 degrees while our algorithm does not impose such a constraint and can efficiently handle any pose, (vi) in order to extract each finger, our algorithm does not use the computationally intensive connected component analysis.

Various features have been used in hand biometrics-based recognition systems. For instance, Gabor filters, line features and Principal Component Analysis (PCA) [8], Zernike moments [12], and the Hausdorff distance and Independent Component Analysis (ICA) [7]. Fusing different biometric modalities (i.e., face, fingerprints, and hand) has shown to improve the performance compared to unimodal biometric systems [10]. However, the fusion of information from different parts of the same biometric has been considered to a lesser extent. The limited work in this area include Kumar and Zhang's work [15] who investigated the feature selection of hand shape and palmprint features. Cheung et al. [19] proposed a two-level fusion strategy for multimodal biometric verification. The Discrete Cosine Transform (DCT) and line features are also extracted for the palmprints. In contrast, we use both shape and texture based features to improve the accuracy and reliability of the recognition results. Our approach uses a two-level fusion strategy for multimodal biometrics. The first level of fusion is based on the separate matching of the scores of the fingers and the palmprints. The second level fuses the scores of the first fusion level to obtain the combined score for the entire hand. Based on this score, the final recognition decision is made.

The rest of the paper is organized as follows: an overview of the proposed system is presented in Section II. Section III describes our hand image segmentation and pose-correction algorithms. Section IV presents the various feature extraction, matching and fusion techniques. Experimental results and analysis are presented in Section V with some concluding remarks drawn in Section VI.

## II. THE OVERALL HAND RECOGNITION SYSTEM

The proposed hand based identification and verification system consists of four major building blocks (see Figure 1): (i) segmentation and pose normalisation of the fingers and the palm, (ii) feature extraction, (iii) feature matching, (iv) fusion and recognition. The first block corresponds to the aforementioned segmentation and pose normalisation module, while the last three blocks correspond to the recognition module.

The operation of the system can be divided into two phases, an offline enrolment phase and an online recognition phase. In both phases, an image of an individual's hand is acquired. The hand is then segmented from the image. The pose of the palm and the fingers are then estimated. The palm and fingers are then cropped and their poses are corrected to poses and sizes that are consistent/standard for all images. Various features from the fingers and the palm are extracted using relevant features extraction algorithms. These features are then stored during the enrolment phase, in a Feature Library, also referred to as the *gallery*. In the recognition phase, the relevant palm/finger features from the image of an individual's hand are extracted. This is the so called *probe*. The features are then compared with the relevant features from the Feature Library and a matching score is obtained. The matching scores are then fused together to obtain the final score. Finally, a decision is made based on the best matching score. A detailed description of the proposed system is presented in Sections III and IV.

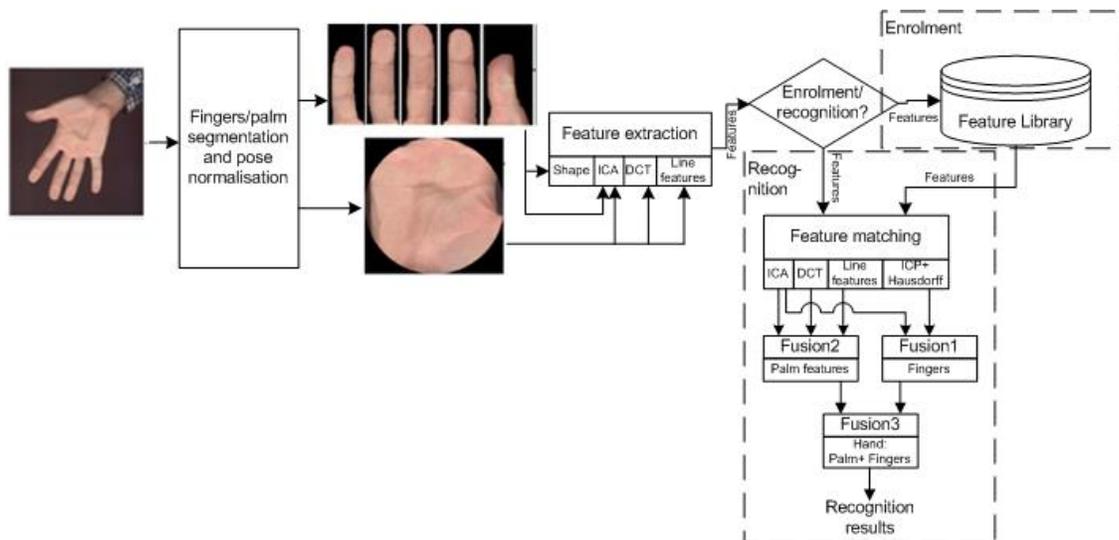

Figure 1: The proposed robust pose invariant shape and texture based hand recognition system.

### III. HAND SEGMENTATION, POSE ESTIMATION AND CORRECTION

A major contribution of this paper is a robust and automatic hand segmentation and pose normalisation system that is able to (i) segment the hand from an image even in the presence of a noisy background, (ii) estimate the global pose of the hand and the individual pose of each finger, (iii) correct the hand and finger poses and re-orient them into poses that are consistent for all images, and finally (iv) segment the palm and fingers, and represent each in a consistent representation that is invariant to hand size. The procedure of this algorithm is briefly explained in the following steps with the aid of an illustrative example shown in Figure 2.

#### A. Proposed Hand Segmentation, Pose Estimation and Correction Algorithm

**Background Segmentation:** The algorithm starts by segmenting the hand from its background and excluding other objects in the case of clutter. For this task, we propose a background segmentation algorithm which performs the following steps: (i) Filter the image from the high frequency noise (small blobs) using a Gaussian kernel. (ii) Generate a binary mask for the foreground (hand) in the following way: the filtered image is converted to a greyscale image and compared against a threshold estimated using the Multi Otsu's method. Otsu's method is well-known to convert a gray level image to a binary one by performing histogram shape-based image thresholding [17].

The method assumes that the gray image contains two classes of pixels (bi-modal histogram), one for the foreground and the other for the background. It then calculates the optimum threshold to separate the two classes such that their combined spread (intra-class variance) is minimal [20], (iii) Correct for potential errors resulting from Otsu's method to improve the segmentation process. This is accomplished using a skin colour detector which employs a histogram-based Bayesian classifier technique [21]. The Bayesian technique was tested in [21] on large databases and was found to have higher classification rates compared with other classifiers including the piecewise linear and Gaussian classifiers. However, in our case, in several images the background was found to have a skin-colour like appearance which affected the performance of the segmentation. Other factors such as the variations of the skin and the background colour from one region to another also affected the performance of the segmentation. In order to overcome these problems and to further improve the performance of the segmentation process, the original image is divided into sub-images (we divided the image into 4 regions) of smaller sizes and the results of Otsu's and the Bayesian skin colour detector are combined to form a binary mask to accurately segment the foreground. Few basic image morphing processes are then applied to remove isolated pixels and to merge large blobs with the dominant foreground (i.e., the largest blob).

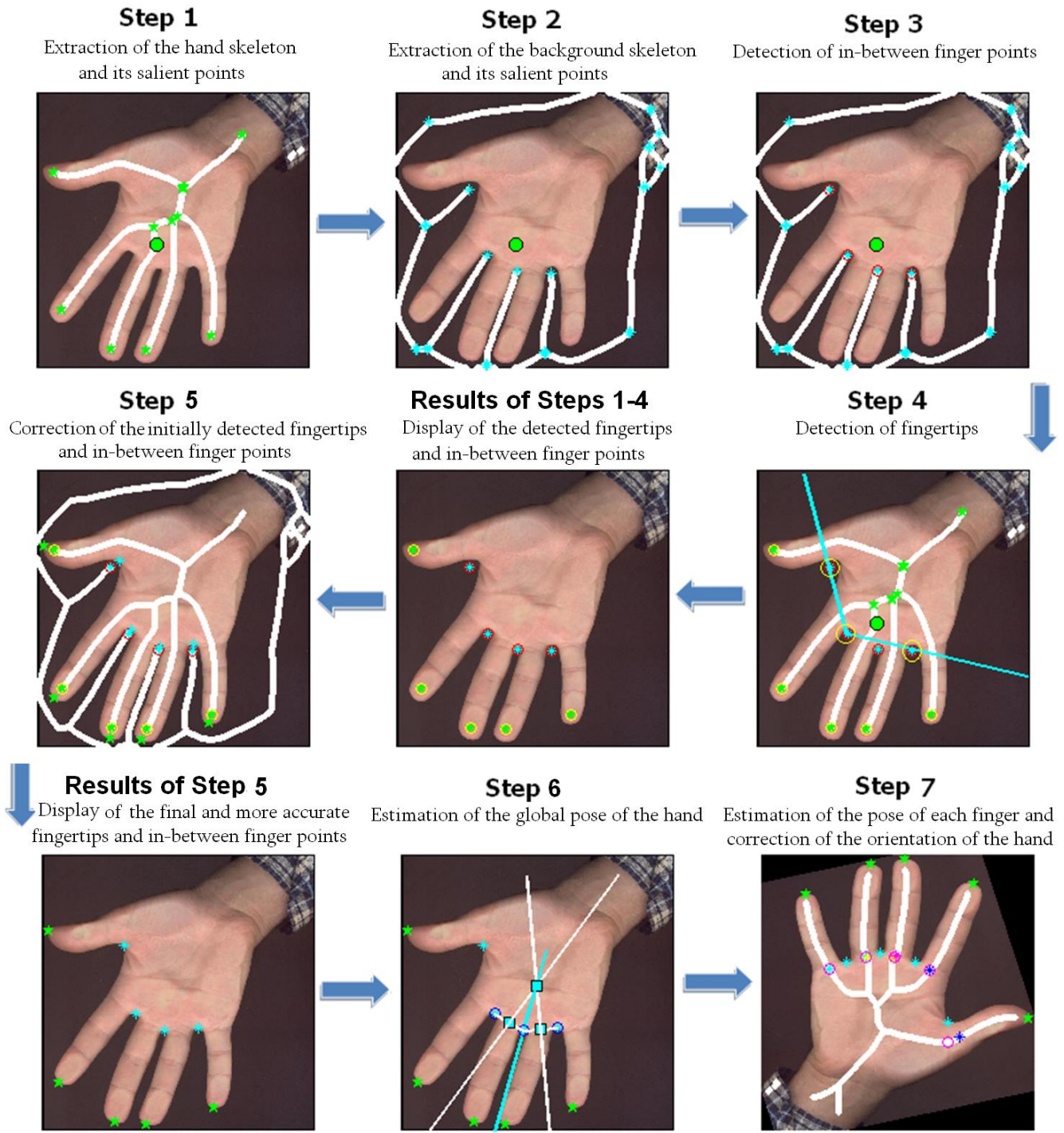

Figure 2: Steps describing the proposed hand's palm and fingers segmentation, pose estimation and correction algorithm (picture best viewed in colour).

**Hand and Fingers Pose Estimation and Correction:** Once the hand is segmented, we estimate the global pose of the hand and the pose of each finger. In order to do that we first determine the locations of a number of salient points of the hand, the fingertips and the in-between finger points (valley points). We then use the in-between finger points to estimate and correct the global pose of the hand, and we finally use both of the fingertips and the in-between finger point locations to estimate the pose of each finger. Our algorithm follows the steps below to locate these points and to correct the hand and finger poses (see Figure 2 for illustrations).

**Step 1: Extraction of the hand skeleton and its salient points:**

First, the binary silhouette image of the segmented hand is extracted, its skeleton is determined using our image skinning technique which shrinks the binary image using an edge detector and a Gaussian point spread function (PSF). Further details can be found in [20]. The Centre of Mass (CoM) of the extracted hand skeleton is then calculated (shown by a green circle in Figure 2, Step 1). Finally, a combined edge-then-corner detection algorithm is then applied to extract the corner points (including the endpoints) of the hand's skeleton (green stars in Figure 2). The corner points are detected based on global and local curvature properties of the edges of the skeleton. These properties include the maximum angle of a potential corner and the minimum ratio of the major to minor axes of an ellipse fitting the edge points around the potential corner [22].

**Step 2: Extraction of the background skeleton and its salient points**

The same approach used to extract the foreground skeleton and its corners in Step 1 is used to extract the skeleton and the corner points of the background. For that purpose, we simply use the negative of the background binary image. Image morphing is then applied to eliminate any noise blobs. The corner points are shown by cyan stars in Figure 2, Step 2.

**Step 3: Detection of in-between finger points**

The in-between finger points (cyan stars with red circles in Figure 2, Step 3) are identified as the four points amongst the corner points which are detected from the background skeleton in Step 2, and have the following criteria: i) They are the nearest to the CoM point, and ii) They are the highest tip of the endpoints of the background skeleton corners.

It has been reported in [23, 24] and supported by several in house experiments, that these in-between-finger points (tips of the fingers valley) are robust to hand, fingers poses and the finger valley size variations, hence can be used as robust salient points for hand pose correction. We developed a repeatable, accurate and automatic approach for the extraction of these points and used them for hand pose corrections.

**Step 4: Detection of fingertips**

To search for the fingertips, three points are identified (larger yellow circles in Figure 2 Step 4) from the four in-between finger points detected in Step 3. This is accomplished as follows. First, two furthest points from the four previously identified in-between finger points are identified. These two points correspond to the point between the thumb and the index finger, and the point between the little and ring fingers. However, they are not identified in any particular order and we need a way to determine which one is which. Since the point between the little and the ring fingers has the shortest distance from one of the two other remaining points, this enables us to correctly identify and uniquely label that point and the point between the ring and middle fingers. Subsequently, we identify and label all of the four in between finger points. We further extend this process to check whether the hand is a right or a left one. This is done by comparing the location of the point close to the thumb with respect to the other three points. If they are in the clockwise order then it is a right hand; otherwise it is a left hand. Finally, to identify the fingertips (green stars with yellow circles) from all other hand skeleton points (green stars), an area of interest (AOI), which contains all fingertips and excludes the arm and other objects (e.g., clothes, watches) attached to it, is defined. Two lines (cyan) connecting the point between the index and the middle fingers and the two furthest points are used. The image side where the two lines make the larger angle represents the AOI for fingertips identification. As shown in Figure 2 Step 4, this AOI excludes the arm and other objects attached to it (e.g., clothes and watches). It therefore, makes our algorithm robust to the presence of clothing and hand accessories.

**Results of Steps 1-4:**

This subfigure shows the identified fingertips and the in-between finger points resulting from the procedure in Steps 1 to 4. Due to the skeletonising operation which includes filtering using a Gaussian PSF of a certain kernel-size (we used 9×9 in this example), these points deviate from their accurate locations by at least a kernel-size of the PSF i.e., 9 pixels. Therefore, the locations of these points need to be refined and corrected.

**Step 5: Correction of the initially identified fingertips and in-between finger points**

To correct the obtained points from the previous step, we extend the skeleton lines from the identified fingertips and the in-between finger points by a kernel size (i.e. at least 9 pixels) along the direction of these points as shown in Fig. 2 Step 5. We then obtain their final locations as the intersection of the respective skeleton lines and the hand contour.

**Results of Step 5:**

This subfigure displays the final fingertips and the in-between finger points locations.

**Step 6: Estimation of the global pose of the hand**

To estimate the global pose of the hand, the three in-between finger points (shown in Figure 2 Step 6) are used as follows. Two lines from the point between the middle and ring fingers and each of the two other neighbouring in-between points (e.g., one point is the point between the ring and the little fingers; while the other is the point between the middle and the index fingers) respectively are drawn and their midpoints are determined. Two perpendicular lines (white) to those lines through their midpoints are then drawn and their intersection is also determined. We define the global pose of the hand by the line (cyan) which passes through the above intersecting point and the point between the middle and ring fingers. The global pose angle of the hand is then calculated from the slope of the line.

**Step 7: Calculation of the pose of each finger and correction of the pose of the hand**

Using the global hand pose angle obtained above, the hand pose/orientation is corrected, of course along with the fingertips and the in-between finger points. Using the rotated in-between finger points, the midpoint of each finger is determined and the pose of each finger is then calculated using the slope of the line which passes through the fingertip and the finger midpoint.

**Final results: Cropping the re-oriented palm and fingers for feature extraction**

Using the detected in-between finger points, fingertips and finger poses, each finger is cropped and reoriented in a vertical pose and resized to a pre-defined image size to ensure consistency for all hands. The hand palm is cropped by fitting a circle to the in-between finger points and the fingers' midpoints. We then and use its radius to create a binary disc/mask to crop the palm. For consistency and to overcome any scaling problem, the cropped palm is also resized to fit a circle of a pre-defined radius.

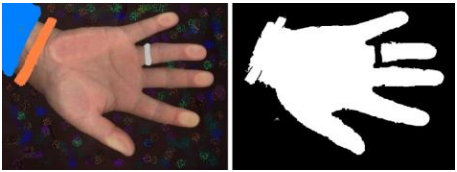

Figure 3a: An illustrative example of background segmentation using the proposed algorithm (note the artificially introduced ring, background noise, clothes, and bracelet).

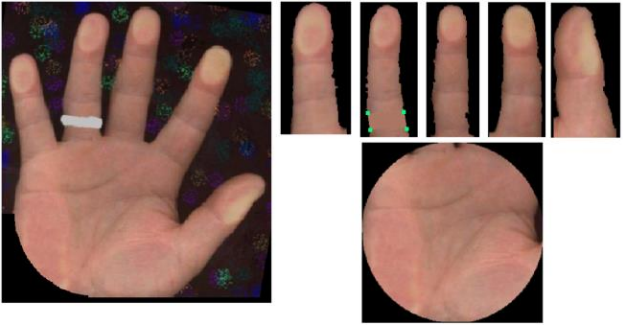

Figure 3b: The segmented and pose corrected hand, palm and fingers.

**An illustrative Example of the Robustness of the Segmentation Algorithm:** To illustrate the robustness of our algorithm to background colour variations and the robustness of the segmentation of the palm in the presence of clutter (other objects), Figures 3a and 3b show the results of a low resolution hand image at nearly 90 degrees from North, with a noisy background, clothing, and a ring. Although the background noise was severe, our combined skin colour detector and the subimages thresholding algorithm were able to accurately segment the hand from the background. Figure 3b shows the entire segmented, pose corrected, hand, palm and fingers. It should be noted that the pose of the fingers and the palm have been reoriented and aligned to a vertical direction (zero degrees from North). An edge detection algorithm is employed to identify the corners of the missing finger parts caused by the presence of the ring. These parts are then filled with a skin colour calculated using the average colour of the segmented finger.

## IV. FEATURE EXTRACTION AND FEATURE MATCHING STRATEGIES

Once the fingers and the palm are segmented from the image and their poses are corrected, various shape and appearance based features are extracted (Figure 4). The palm and each finger of the hand are considered separately for feature extraction. Shape (ICP) and appearance (ICA) features are extracted from each finger. The Iterative Closest Point (ICP) algorithm [25] is employed to align a probe finger with a gallery finger and their geometric distance is measured using the Hausdorff distance which represents a similarity score between the finger pair. For appearance, the ICA algorithm provides statistically independent features. The minimum Euclidean distance between the probe and gallery appearance features is used as a similarity score. These steps are employed separately for all five fingers. The scores of all fingers are then fused together in Fusion1 Step using a weighted sum (see Section IV-E, eq. (9)) approach. In this Step, the similarity score of the fingers of a hand is also obtained using the *max* rule so that the algorithm picks up the best matched fingers-set.

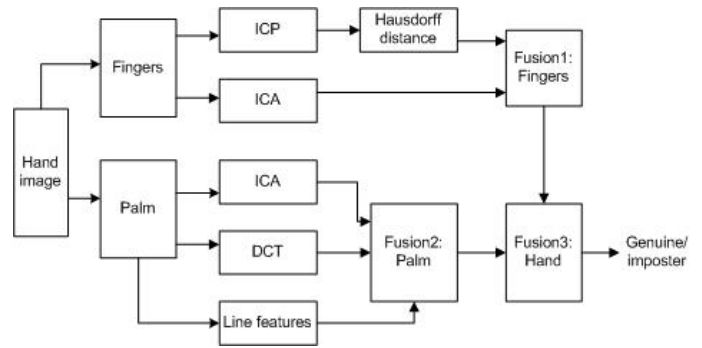

Figure 4: A block diagram of the feature extraction and feature matching techniques of the proposed system.

For the palm, we only extract appearance and texture features. Shape features are not considered since the shape of the palm is affected by the orientation of the thumb and can partially be occluded (e.g., with clothing). Texture features using the DCT and directed-line-feature methods are extracted and similarity scores are fused together in Fusion2 using the *max* rule. For appearance, the ICA algorithm is used with the minimum Euclidean distance between probe and gallery palms. Finally, the scores of Fusion1 and Fusion2 are fused together using the *mean* rule to obtain the final score of the hand recognition (Fusion3). It should be noted that for multimodal biometrics it has been shown in [26] that a simple mean rule performs better than other classification schemes such as a decision tree and a Linear Discriminant Analysis (LDA) coupled with a minimum distance rule. A brief description of the various feature extraction, matching, and fusion techniques that are used in this paper is provided in Figure 4.

*A. The iterative closest point (ICP) algorithm*

The ICP algorithm aims to find the transformation (translation and rotation) between two pointclouds (e.g., a pointcloud and a reference surface), by minimizing the square errors between them [23]. It can be used in 2D or 3D and in our case the pointclouds correspond to the 2D shapes of the fingers we are trying to match. The matching is performed using a similarity measure or distance (in our case the Hausdorff distance). ICP iteratively revises the transformation needed to minimize the distance between the two shapes. The algorithm has four major steps: (i) associate the points between the two shapes, (ii) estimate the transformation parameters using a mean-square cost function, (iii) transform the points using the estimated parameters, and (iv) iterate the previous steps until the stopping criteria is met.

*B. The modified Hausdorff distance*

A number of geometric distances/metrics are proposed in the literature [27]. The Hausdorff distance metric is one of the most popular metrics which has been used to estimate the similarity between different shape geometries. This metric has been extensively used for binary image comparison and computer vision [7]. The advantage of the Hausdorff distance, over binary correlation for example, is the fact that the Hausdorff distance measures the proximity rather than the exact superposition of the two shapes. It is therefore more tolerant to perturbations/variations in the locations of the points. In addition, since the original definition of the Hausdorff distance has been shown to be sensitive to local noise, we opt to use a more robust version of this metric, namely the modified Hausdorff distance [28]. For the sake of completeness, a brief description of this distance is now presented. Given two sets $S$ and $T$ of the contour pixels of two shapes (e.g., fingers), represented by the sets $S = \{s_1, s_2, \cdots, s_{N_s}\}$, $T = \{t_1, t_2, \cdots, t_{N_t}\}$, where $\{s_i\}$ and $\{t_j\}$ denote contour pixels in the Cartesian coordinate system for $i = 1, \cdots, N_s$ and $j = 1, \cdots, N_t$, the modified Hausdorff distance [28] is defined as follows:

$$H_{hausdorff}(S,T) = \max(h(S,T), h(T,S)) \quad (1)$$

where
$$h(S,T) = \frac{1}{N_s} \sum_{s \in S} \min_{t \in T} \|s - t\| \quad (2)$$

$$h(T,S) = \frac{1}{N_t} \sum_{t \in T} \min_{s \in S} \|s - t\| \quad (3)$$

$\|s - t\|$ is a norm over the elements of the two sets and obviously the contour pixels $(s,t)$ run over the set of indices $i = 1, \cdots, N_s$ and $j = 1, \cdots, N_t$. In our case this norm is taken to be the Euclidean distance between the two points.

*C. Independent Component Analysis*

ICA is another widely used technique for feature extraction. We employ ICA to extract statistically independent variables from a mixture of variables. It has been successfully used in many applications to find hidden factors within data to be analyzed or to decompose it into the original source signals. In this paper, ICA is applied on grayscale images to extract and summarize prototypical shape information. ICA has several advantages – it is simple to calculate and it explicitly takes advantage of the image statistics as explained below [29]. ICA assumes that each observed signal $\{x_i(k)\}, k = 1, \cdots, K$ is a mixture of a set of $N$ unknown independent source signals $l_i$, through an unknown mixing matrix A. With $x_i$ and $l_i$ forming the rows of the N×K matrices X and L, respectively, the following model is obtained:

$$X = AL. \quad (4)$$

The data vectors for the ICA analysis are the lexicographically ordered image pixels. The dimension of these vectors is K (for example, K = 30,720, we used (160×192) images). ICA aims to find a linear transformation W for the inputs which minimize the statistical dependence between the output component $y_i$, the latter being estimates of the hypothesized independent sources $l_i$:

$$\widehat{L} = Y = WX \quad (5)$$

Such a transformation W, is commonly referred to as the de-mixing matrix. For the purpose of this research, the fast ICA algorithm [29] was implemented in order to find W. There exist two architectures of ICA [29] – $ICA_1$ and $ICA_2$, respectively assume the basis images or their mixing coefficients to be independent. $ICA_2$ architecture produces global features in the sense that every image feature is influenced by every pixel. Depending on the preference, this makes them either susceptible to occlusions and local distortions, or sensitive to holistic properties. Alternatively, $ICA_1$ produces spatially localized features that are only influenced by small parts of the image. While the details of both ICA techniques are available in [7, 29], for the sake of completeness a brief overview is included here. It should be noted that in our experiments we found that $ICA_2$ consistently provided better identification and verification results compared to $ICA_1$. Therefore, only the $ICA_2$ technique is described here.

**$ICA_2$**: In this architecture, the superposition coefficients are assumed to be independent. Thus, this model assumes each of $K$ pixels of the hand images to result from independent mixtures of random variables, that is the "pixel sources". For this purpose, the transpose of the data matrix: $X^T$ is considered. However, the huge dimensionality of pixel vectors (typically $K \gg N$) necessitates a PCA reduction stage prior to ICA, in order to obtain $M$ ($M \leq N$) principal components. In fact, the eigenvectors of the $K \times K$ covariance matrix $X^T X$ can be calculated using the eigenvectors of the much smaller $N \times N$ matrix $XX^T$. The orthonormal eigenvectors of $X^T X$ matrix can be calculated from the eigenvectors of the $XX^T$ matrix using the singular value decomposition (SVD) theorem [30].

Let $\{v_1, v_2, \cdots, v_M\}$ be the $M$ ranked eigenvectors with eigenvalues $\{\lambda_1 \geq \lambda_2 \geq \cdots \geq \lambda_M\}$ of $XX^T$ matrix. Then, by SVD theorem, the corresponding orthonormal eigenvectors of $X^T X$ matrix are $\{u_1, u_2, \cdots, u_M\}$ where $u_j = \frac{1}{\sqrt{\lambda_j}} X v_j$. After the projection of the input vector $x$ onto the eigenvectors $u_j$, we obtain the $j$th feature $y_j = \frac{1}{\sqrt{\lambda_j}} v_j^T X X^T = R X^T$, where $R$ is called the projection operator. The hand image data is then reduced after being projected onto the $M$ principal components. Finally, we decompose $R X^T$ to source and mixing coefficients. We obtain the basis functions (the hand images) in the columns of the estimated mixing matrix A ($N \times N$). Consequently, the coefficients in the estimated source matrix are statistically independent.

In the recognition stage, assuming again that the test hands follow the same model, they are also size reduced with $Rx_{test}^T$, and multiplied by the de-mixing matrix $W = A^{-1}$. The resulting coefficient vector of a test hand $\mathbf{x}_{test}$ ($K \times 1$), found as $\hat{p}_{test} = WRX^T$, which is then compared with the predetermined feature vectors of the training stage. Finally, the individual subject to be tested is simply recognized as the person $i^*$ with the closest feature vector $\hat{p}_{i*}$, where the distance is measured in terms of the cosine of the angle between them:

$$H_{ICA2}(S_i, T) = \frac{\hat{p}_i \bullet \hat{p}_{test}}{\|\hat{p}_i\| \|\hat{p}_{test}\|} \quad (6)$$

*D. Discrete Cosine Transform*

The DCT is one of the most popular transforms in image processing and has been used for various purposes including feature extraction, and recognition [15]. The palmprint image is divided into overlapping blocks of size 16×16 with 4×4 overlaps. The DCT coefficients for each of these blocks are computed. Several of these DCT coefficients have values close to zero and can hence be discarded. It was reported in [8] that the first 12.5% DCT coefficients cover most information to represent the texture information in a hand image and that this information was able to provide a reliable recognition rate. For this reason, only the most significant first 12.5% coefficients are used in our case. The feature vector from every palmprint image is formed by computing the standard deviation of these significant DCT coefficients in each of these blocks. Feature level fusion is employed to obtain the score and the similarity measure between $v_1$ (feature vector from the user) and $v_2$ (stored identity) is used as the matching score and is computed as follows:

$$similarity = \frac{\sum v_1 \cdot v_2}{\sqrt{\sum v_1 \sum v_2}} \quad (7)$$

The similarity measure defined in equation (7) computes the normalized correlation between the feature vector $v_1$ and $v_2$.

*E. Line features*

Palmprint identification using line features has been reported to be powerful and offers high accuracy [31]. The palmprint pattern is mainly made up of palm lines, *i.e.*, principal lines and creases. Line feature matching is reported to be powerful and offers high accuracy in palmprint verification. However, it is very difficult to accurately characterize these palm lines, due to the variations in their magnitudes and directions in noisy images. Therefore, similar to [29], a robust but simple directed-line-feature method is used here.

A robust yet simple directed-line-feature method is proposed in this paper. The segmented palmprint images are normalized so that they have pre-specified mean and variance values (here 100 and 100 respectively). The normalization is used to reduce the possible imperfections in the image due to sensor noise and non-uniform illumination using the method proposed in [32]. For the extraction of line features we use the techniques described in [31]. Four directional line detectors are used to probe the palmprint creases and lines oriented in each of the four directions, i.e. $0^0$, $45^0$, $90^0$, and $135^0$. The spatial extent of these masks was empirically fixed to 5×5. The resultant four images are combined by voting on the gray-level magnitude from corresponding pixel position. The combined image represents the combined directional map of palm-lines and creases in the palmprint image. This image is further divided into several overlapping square blocks. The standard deviation of the grey-level in each of the overlapping blocks is used to form the feature vector for every palmprint image. Feature level fusion is then employed to obtain the similarity score using (7).

*F. Score Level Fusion of the Fingers*

The weighted sum rule has been extensively investigated in the literature and it is the most straightforward fusion strategy at the score level [33]. In this case, the matching scores between pairs of fingers following the query and the template hands are combined into a single score using a weighted sum as follows:

$$score(Q,T) = \sum_{i=1}^{5} w_i \times score(Q_i, T_i) \quad (8)$$

where *score* correspondence to the similarity measure (e.g., distance) between the query $Q$ and the training data $T$. $Q_i$, and $T_i$ represent the $i$th parts of the query and training data.

In our system, the five parts correspond to the little, ring, middle and index fingers, and the thumb. The parameters $w_i$ are the weights associated with the $i$ th part of the hand which needs to satisfy the following constraint:

$$\sum_{i=1}^{5} w_i = 1 \qquad (9)$$

This strategy was originally used in [34] which concluded that the best combination of the weights is as follows: $w_1$ = 0.5/11 (little finger), $w_2$ = 2.5/11 (ring finger), $w_3$ = 3.0/11 (middle finger), $w_4$ = 4.5/11 (index finger), and $w_5$ = 0.5/11 (thumb).

## V. EXPERIMENTAL RESULTS AND ANALYSIS

Experiments were conducted on the dataset from Bogazici University [7]. This hand database contains 1680 colour images of the right hand of 560 different persons, three images of each hand. The images were acquired with a HP Scanjet 5300c scanner at 45-dpi resolution. Each raw image had the spatial resolution of 383×526 pixels prior to the preprocessing stage. A detailed description of the dataset is available in [7].

First, our segmentation and pose correction algorithms were applied to normalise the hand, fingers and the palm. Secondly, the hand identification experiments, based on the normalized hand images were performed on six selected population sizes, namely, population subsets consisting of 20, 50, 100, 200, 500, 560 individuals. The use of different population sizes helps to compare the identification performance with respect to an increasing size of the gallery (number of individuals). A boosting algorithm was applied so that several different formations of subsets (of sizes of 50, 100, 500, and 560) were created by random choice and their performance scores were averaged.

*Identification results:* The identification results are shown in Table 1. We aimed to investigate three different aspects of the identification rate: the effects of fusion, the effects of the number of samples per subject used in training, and the effects of the number of subjects in the training dataset. The first set of experiments was conducted to investigate the independent performance of fingers (Fusion1) and palmprints (Fusion2) based techniques and then the impact of their fusion (Fusion3). The corresponding identification results are presented in Table 1. From the results in Table 1, it is evident that the fusion of fingers and palm improves the identification rates. For instance, as shown in Table 1, with 20 persons, Fusion1 (using only the fingers) and Fusion2 (using only the palm) produced identification rates of 99.20% and 99.10% respectively, when used individually. However, the performance reached 100% when the fusion strategy (Fusion3) was used. Fusion3 consistently provided superior results compared with both Fusion1 and Fusion2 which illustrates the superiority of multimodal versus unimodal biometrics. It is noteworthy to mention that for the fingers both ICA and ICP+Hasudorff distance techniques produced almost equal similarity scores and the resulting identification rate was 99.1% in each case for a sample size of 20.

Table 1: Identification performance (best results are in bold)

|  | Correct identification percentage | | | | | |
|---|---|---|---|---|---|---|
| Enrolment size → | 20 | 50 | 100 | 200 | 500 | 560 |
| Fusion 1 (fingers only) | 99.2 | 99.1 | 98.3 | 98.1 | 96.7 | 96.2 |
| Fusion 2 (palm only) | 99.1 | 99.0 | 98.1 | 97.8 | 96.8 | 96.1 |
| Fusion 3 (hand: palm + fingers) | **100** | **99.5** | **98.9** | **98.7** | **98.3** | **98.2** |

Table 2: Effect of the number of samples (single – S or double – D) per subject used in training on the identification performance (best results are in bold)

|  | Correct identification percentage | | | | | | | | | |
|---|---|---|---|---|---|---|---|---|---|---|
| Enrolment size → | 20 | | 50 | | 100 | | 200 | | 560 | |
| Single or double training data per subject → | S | D | S | D | S | D | S | D | S | D |
| Fusion 1 (fingers only) | 97.2 | **99.2** | 96.1 | **99.1** | 95.9 | **98.3** | 95.3 | **98.1** | 93.2 | **96.2** |
| Fusion 2 (palm only) | 98.4 | **99.1** | 97.9 | **99.0** | 97.6 | **98.1** | 95.6 | **97.8** | 93.7 | **96.1** |
| Fusion 3 (hand: palm + fingers) | 98.8 | **100** | 98.1 | **99.5** | 97.1 | **98.9** | 96.8 | **98.7** | 96.3 | **98.2** |

Table 3: Verification performance as a function of enrolment size (equal error rate)

|  | Correct verification rate | | | | | |
|---|---|---|---|---|---|---|
| Enrolment size → | 20 | 50 | 100 | 200 | 500 | 560 |
| Fusion 1 (fingers only) | 99.9 | 99.3 | 98.7 | 98.2 | 98.0 | 97.9 |
| Fusion 2 (palm only) | 99.7 | 99.2 | 98.6 | 98.1 | 97.8 | 97.8 |
| Fusion 3 (hand: palm + fingers) | **100** | **99.7** | **99.4** | **99.0** | **98.5** | **98.5** |

Secondly, we explored the effect of the number of samples per subject used during training, that is, the impact of multiple independent enrolments of the hand of the same person. We ran the identification experiments with a single training and then with a double training set, both in a round robin fashion. More explicitly, let the three sets of the hand images are referred to as sets A, B, and C. In the single set experiments, the ordering of the test and training sets were {(A,B), (B,A), (A,C), (C,A), (B,C), (C,B)}. In other words, set A hands were tested against the training set of sets B and C separately. In the double training set, the ordering of the test and training sets were {(A, BC), (B, AC), (C, AB)}, e.g., hands in the test set A were recognized using hands in both sets B and C. Finally, the identification scores were averaged from these training and test set combinations. Table 2 indicates significant improvements when the double training datasets were used compared to the single datasets. For instance, with a database of 560 subjects, Fusions 1, 2, and 3 respectively produced average identification rates of (93.2%, 96.2%), (93.7%, 96.1%) and (96.3%, 98.2%) pair-wise for a single and double training set. Therefore, the gains in the identification rate were respectively 3.2%, 2.6%, and 1.97% for Fusions 1, 2, and 3 when two training datasets were used instead of one.

Thirdly, we explored the effect of the number of subjects in the training dataset. Without loss of generality and as expected, a larger dataset produced a lower identification rate. For instance, as shown in Table 1 with Fusion3, the identification rates were 100%, 99.5%, 98.9%, and 98.7% respectively for the database size of 20, 50, 100 and 200 subjects. With a database size of 560 subjects, the identification rate was 98.2% using Fusion3.

*Verification Results:* The next set of experiments was conducted to investigate the verification performance of the proposed technique. During verification, *genuine scores* have to be differentiated from *impostor scores (non-match scores)*. We calculated the distances between the finger/palm shape of the probe and the finger/palm shapes collected in the database (gallery) of the subject that s/he claims to be, and then compared this score against a threshold. If this distance is below the set threshold then the claimant is accepted as true; otherwise s/he is rejected (impostor). In the case of an impostor where the distance to the claimed hand is below the set threshold, then we have a false acceptance. Conversely, if the distance between the probe hand and the gallery is above the threshold we have a case of false rejection. Both false acceptance and false rejection correspond to failures of the verification process.

Figure 5 shows the (line-log) receiver operating characteristic (ROC) curve for Fusion3. At $10^{-4}$ false acceptance rate, our algorithm secured a correct acceptance rate of 87.3%. The verification comparisons between the three feature modalities are given in Table 3 as a function of enrolment size. Note that for smaller populations (sizes 20, 50, 100, and 200), the performance was calculated as the average of several randomly selected subject sets.

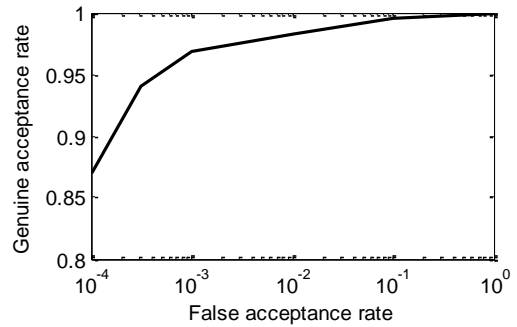

Figure 5: ROC curve for Fusion3.

Table 4: Comparison of the identification performance of algorithms with respect to given enrolment sizes

| Enrolment size | Algorithm in [7] | Proposed algorithm |
|---|---|---|
| 20 | 99.48 | **100** |
| 35 | 99.40 | **99.6** |
| 70 | 99.03 | **99.2** |
| 458 | 97.31 | **98.25** |
| 560 | - | **98.2** |

Tables 4 and 5 summarise comparative results (respectively for the identification and verification results) of the proposed algorithm compared with its counterpart proposed in [7]. The reason for selecting the algorithm in [7] is twofold: First, it has the best performance among all of the existing shape-based hand identification algorithms that are available in the literature. Second, it was tested on the same dataset.

Table 5: Comparison of the verification performance of algorithms with respect to given enrolment sizes

| Enrolment size | Algorithm in [7] | Proposed algorithm |
|---|---|---|
| 20 | 99.55 | **100** |
| 50 | 99.40 | **99.7** |
| 100 | 98.85 | **99.4** |
| 458 | 98.21 | **98.55** |
| 560 | - | **98.5** |

It should be noted that our proposed system uses the fusion of hand shape and palmprint texture data. On the other hand, the work in [7] only used shape-based features of the fingers and the palm. To justify the strength of the fusion of shape and texture in our proposed system, we further compare the results with palm-print only recognition. For instance, the multiple (two) palmprint based method proposed in [8] achieved a Genuine Acceptance Rate (GAR) of 91% (out of only 100 subjects) at the False Acceptance Rate (FAR) of $10^{-2}$. In contrast our proposed system achieved more than 97% GAR at the same FAR (see Figure 5). Therefore, the experimental results above show the superiority of the proposed system compared to other existing systems in terms of accuracy rate and the size of the dataset.

## VI. CONCLUSIONS AND FUTURE WORK

In this paper, we presented a new system for hand biometric based identification and verification. The proposed system has several advantages over existing ones. It is peg-free and fully automatic and is able to segment the palm and fingers from the image and exclude "unwanted" objects (clutter). Regardless of the pose of the hand and the poses of each finger in an image, the system is able to estimate and correct the poses of the hand and fingers. It extracts a consistent scale invariant representation of the palm and each finger. The proposed system achieved consistently better results compared to state of the art algorithms. The superior results are due to the combination of a new robust pose invariant hand segmentation algorithm followed by efficient feature extraction, feature matching and fusion techniques. Extensions include prior processing to distinguish the hand of a male from the hand of a female to reduce the matching search space for a faster recognition. We also aim to target hand images in which one or more fingers are joined to the adjacent one (no gap between fingers) and to cases where a person has fewer or more than five fingers (polydactylism).

## ACKNOWLEDGMENT

This work was tested on the hand image database provided by Prof Sankur of Bogazici University.

## REFERENCES

[1] A. K. Jain, L. Hong, S. Pankanti, and R. Bolle, "An identity authentication system using fingerprints," *The Proceedings of IEEE,* vol. 85 no. 9, pp. 1365-1388, 1997.
[2] A. S. Mian, M. Bennamoun, and R. Owens, "An efficient multimodal 2D-3D hybrid approach to automatic face recognition," *IEEE Transactions on Pattern Analysis and Machine Intelligence* vol. 29, no. 11, pp. 1927-1943, 2007.
[3] A. S. Mian, M. Bennamoun, and R. Owens, "Keypoint detection and local feature matching for textured 3D face recognition," *International Journal of Computer Vision (IJCV),* vol. 79, no. 1, pp. 11-12, 2008.
[4] S. Islam, M. Bennamoun, R. Owens, and R. Davies, "A review of recent advances in 3D ear and expression invariant face biometrics," *ACM Computing Surveys (in press),* vol. 44, no. 4, 2012.
[5] K. W. Bowyer, K. Hollingsworth, and P. J. Flynn, "Image understanding for iris biometrics: A survey " *Computer Vision and Image Understanding,* vol. 110, no. 2, pp. 281-307, 2008.
[6] A. El-Sallam, F. Sohel, and M. Bennamoun, "Robust pose invariant shape-based hand recognition," presented at the 6th IEEE Conference on Industrial Electronics and Applications (ICIEA), 2011.
[7] E. Yoruk, E. Konukoglu, B. Sankur, and J. Darbon, "Shape-based hand recognition," *IEEE Transactions on Image Processing,* vol. 15, no. 7, pp. 1803-1815, 2006.
[8] A. Kumar and D. Zhang, "Personal authentication using multiple palmprint representation," *Pattern Recognition,* vol. 38, no. 3, pp. 1695 – 1704, 2005.
[9] A. K. Jain, A. Ross, and S. Prabhakar, "An introduction to biometric recognition," *IEEE Transactions on Circuits and Systems for Video Technology,* vol. 14, no. 1, pp. 4-20, 2004.
[10] A. Jain, K. Nandakumar, and A. Ross, "Score normalization in multimodal biometric systems," *Pattern Recognition,* vol. 38, pp. 2270 – 2285, 2005.
[11] R. Sanchez-Reillo, C. Sanchez-Avila, and A. Gonzalez-Marcos, "Biometric identification through hand geometry measurements," *IEEE Transactions on Pattern Analysis and Machine Intelligence,* vol. 22, no. 10, pp. 1168–1171, 2000.
[12] G. Amayeh, G. Bebis, A. Erol, and M. Nicolescu, "Hand-based verification and identification using palm–finger segmentation and fusion," *Computer Vision and Image Understanding,* vol. 113 pp. 477–501, 2009.
[13] A. K. Jain, A. Ross, and S. Pankanti, "A prototype hand geometry based verification system," presented at the 2nd Int. Conf. Audio- and Video-Based Biometric Person Authentication, 1999.
[14] A. K. Jain, A. Ross, and S. Prabhakar, "Biometrics-BasedWeb Access," 1998.
[15] A. Kumar and D. Zhang, "Personal recognition using hand shape and texture," *IEEE Transactions on Image Processing,* vol. 15, no. 8, pp. 2454–2461, 2006.
[16] D. Z. a. A. Kumar, "Personal Recognition Using Hand Shape and Texture," *IEEE Transactions of Image Processing,* vol. 15, no. 8, pp. 2454-2461, 2006.
[17] C. Öden, A. Erçil, and B. Büke, "Combining implicit polynomials and geometric features for hand recognition," *Pattern Recognition Letters,* vol. 24, pp. 2145–2152, 2003.
[18] H. D. E. Yoruk, and B. Sankur, "Hand Biometrics," *Image and Vision Computing,* vol. 24, no. 5, pp. 483-497, 2006.
[19] M. Cheung, M. Mak, and S. Kung, "A two level fusion approach to multimodal biometrics verification," presented at the IEEE International Conference on Acoustics, Speech, and Signal Processing (ICASSP'05), 2005.
[20] P.-S. Liao, T.-S. Chen, and P.-C. Chung, "A fast algorithm for multilevel thresholding," *Journal of Information Science and Enginerring,* vol. 17, no. 5, pp. 713-727, 2001.
[21] S. L. Phung, A. Bouzerdoum, and D. Chai, "Skin segmentation using color pixel classification: analysis and comparison," *IEEE Transactions on Pattern Analysis and Machine Intelligence,* vol. 27, no. 1, pp. 148-154, 2005.
[22] X. C. He and N. H. C. Yung, "Corner detector based on global and local curvature properties," *Optical Engineering,* vol. 47, no. 5, 2008.
[23] S. Ribaric and I. Fratric, "A biometric identification system based on eigenpalm and eigenfinger features," *IEEE Transactions on Pattern Analysis and Machine Intelligence,* vol. 27, no. 11, 2005.
[24] L. Fang, M. Leung, and C. Chian, "Making palm print matching mobile," *International Journal of Computer Science and Information Security (IJCSIS),* vol. 6, no. 2, pp. 1-9, 2009.
[25] P. J. Besl and N. D. McKay, "A method for registration of 3-D shapes," *IEEE Trans. Pattern Anal. Mach. Intell,* vol. 14, pp. 239–256, 1992.
[26] S. Garcia-Saliccetti, M. A. Mellakh, L. Allano, and B. Dorizzi, "Multimodal biometric score fusion: the mean rule vs. support vector classifiers," presented at the 13th European Signal Processing Conference (EUSIPCO), 2005.
[27] F. A. Sohel, G. C. Karmakar, L. S. Dooley, and M. Bennamoun, "Geometric distortion measurement for shape coding: a contemporary review," *ACM Computing Surveys,* vol. 43, no. 4, 2011.
[28] B. Takacs, "Comparing face images using the modified Hausdorff distance," *Pattern Recognition,* vol. 31, no. 12, pp. 1873-1881, 1998.
[29] A. Hyvarinen and E. Oja, "Independent component analysis: Algorithms and applications," *Neural Networks,* vol. 13, no. 4-5, pp. 411-430, 2000.
[30] G. H. Golub and C. F. van-Loan, *Matrix Computation*, 3rd ed.: The John Hopkins University Press, Baltimore, 1996.
[31] A. Kumar, D. C. Wong, H. C. Shen, and A. K. Jain, "Personal verification using palmprint and hand geometry biometric," in *Time-Varying Image Processing and Moving Object Recognition*, ed Guildford, UK: Springer, 2003, pp. 668–678.
[32] L. Hong, Y. Wan, and A. K. Jain, "Fingerprint image enhancement : Algorithm and performance evaluation," *IEEE Transactions on Pattern Analysis and Machine Intelligence,* vol. 20, pp. 777-789, 1998.
[33] J. Kittler, M. Hatef, R. Duin, and J. Matas, "On combining classifiers," *IEEE Transactions on Pattern Analysis and Machine Intelligence,* vol. 20, no. 3, pp. 226-139, 1998.
[34] G. Amayeh, G. Bebis, A. Erol, and M. Nicolescu, "A component-based approach to hand verification," presented at the IEEE Conference on Computer Vision and Pattern Recognition (CVPR '07), 2007.